%% file: pellier20e.tex
\def\year{2020}\relax
\documentclass[letterpaper]{article} % DO NOT CHANGE THIS
\usepackage{aaai20}  % DO NOT CHANGE THIS
\usepackage{times}  % DO NOT CHANGE THIS
\usepackage{helvet} % DO NOT CHANGE THIS
\usepackage{courier}  % DO NOT CHANGE THIS
\usepackage[hyphens]{url}  % DO NOT CHANGE THIS
\usepackage{graphicx} % DO NOT CHANGE THIS
\usepackage{graphicx}  % DO NOT CHANGE THIS
\usepackage[ruled,linesnumbered]{algorithm2e}
\usepackage[hyphens]{url}
\usepackage{multirow}
\usepackage{subcaption}
\usepackage{amsmath,amssymb,amsthm}
\usepackage{bbm}
\usepackage[ruled,linesnumbered]{algorithm2e}
\usepackage{subfiles}

\title{AMLSI: A Novel and Accurate Action Model Learning Algorithm}
\author{
Maxence Grand, Damien Pellier, Humbert Fiorino\\
Univ. Grenoble Alpes, LIG\\
3800 Grenoble, France\\
\{Maxence.Grand, Damien.Pellier, Humbert.Fiorino\}@univ-grenoble-alpes.fr
}

\begin{document}
\maketitle
\input{parts/abstract.tex}
\input{parts/intro.tex}

\input{parts/related.tex}
\input{parts/problem.tex}
\input{parts/domain.tex}
\input{parts/experiment.tex}
\input{parts/conclusion.tex}
\input{parts/ack.tex}
\bibliography{biblio}
\bibliographystyle{aaai}
\end{document}

%% file: parts/abstract.tex
\begin{abstract}

This paper presents new approach based on grammar induction called AMLSI (\textbf{A}ction \textbf{M}odel \textbf{L}earning with \textbf{S}tate machine \textbf{I}nteractions). The AMLSI approach does  not require a training dataset of plan traces to work. AMLSI proceeds by trial and error: it queries the system to learn with randomly generated action sequences, and it observes the state transitions of the system, then AMLSI returns a PDDL domain corresponding to the system. A key issue for domain learning is the ability to plan with the learned domains. It often happens that a small learning error leads to a domain that is unusable for planning. Unlike other algorithms, we show that AMLSI is able to lift this lock by learning domains from partial and noisy observations with sufficient accuracy to allow planners to solve new problems.
\end{abstract}

%% file: parts/intro.tex
\section{Introduction}
\label{sec:intro}

Many real world systems implicitly rely on state machines. In communicating systems, for example, each party has to follow the same communication protocol or the system could deadlock. Each party follows a state machine where an action like sending or receiving a message puts the overall system into a new state. For instance, an ATM dispenses cash only when the machine is in a state where a card has been inserted and the PIN verified.

Planning Domain Definition Language (PDDL) \cite{pddl} allows to model state machines and to plan action sequences achieving targeted goals. It is generally accepted that hand-encoding PDDL is difficult, tedious and error-prone. The experts of the system to model are not always PDDL experts and vice versa. Planning domain learning algorithms have been proposed to automatically generate PDDL domains \cite{arora2018review}. A challenging issue for these learning algorithms is their ability to generate domains that planners can use to solve new planning problems. In practice, most state of the art approaches does not evaluate this ability. They use the syntactical error (differences in the preconditions and the effects of operators) between International Planning Competition (IPC) benchmarks and learned domains. Unfortunately, it is not because one learned domain is syntactically closer to an IPC domain that it is better than another. It often happens that a small syntactical error leads to a domain that is unusable for planning: the ability of a PDDL domain to solve planning problems can depend on a few number of decisive preconditions/effects.

In this paper, we propose a new approach based on grammar induction called AMLSI (\textbf{A}ction \textbf{M}odel \textbf{L}earning with \textbf{S}tate machine \textbf{I}nteractions), allowing to "retro-engineer" real world state machines as accurate PDDL domains. The AMLSI approach does  not require a training dataset of plan traces to work. AMLSI proceeds by trial and error: it queries the system to learn with randomly generated action sequences, and it observes the state transitions of the system, then AMLSI returns a PDDL domain corresponding to the system. For instance, in the ATM example, sequences of actions like inserting a card, typing a number, aborting money withdrawal etc. are tested. These action sequences can possibly be infeasible. No prior knowledge on correct sequences is required. Unlike other approaches, we show that AMLSI is able to learn domains from partial and noisy observations with sufficient accuracy to allow planners to solve new problems.

The rest of the paper is organized as follows. In section \ref{sec:related} we present the related works. In section \ref{sec:problem} we propose a problem statement and, in section \ref{sec:amlsi}, we detail the AMLSI algorithm. Finally, section \ref{sec:exp} evaluates the performance of AMLSI on IPC benchmarks. Also, AMLSI's performance is compared with the LSO-NIO \cite{lsonio} algorithm which is the approach with the closest input setting. Indeed, LSO-NIO takes as input random walks including action failures with partial and noisy observations.

%% file: parts/related.tex
\section{Related Works}\label{sec:related}

Many approaches have been proposed to learn planning domains. These works can be classified according to the input data of the learning process and the complexity of the language used in the output domain. The input data can be plan traces obtained by resolving a set of planning problems, partial planning domains to complete or random walks. The input data can be complete (states and actions), partial, completely blind,or noisy. As output, the learned planning domains can have different levels of expressivity: negative preconditions, static relations, numerical functions or conditional effects.

Many works takes as input a set of plan traces and a partial action model, and tries to incrementally refine this action model to complete it, as for instance, EXPO \cite{expo} and more recently RIM \cite{rim} and OpMaker \cite{opmaker}. In practice, RIM constructs sets of "soft" and "hard" constraints between observed states and actions, which are solved with weighted MAX-SAT solvers to obtain the refined action models. In all these approaches, it is assumed that the observations are complete and noiseless. Opmaker induces operators with users interaction. Th partial domain is built by the user within the GIPO tools \cite{gipo}, then users gives plan traces and intermediate state observations and OpMaker induces pre-conditions and effects.

A second group of works takes as input only plan traces. Most of them are able to deal with partial observations (except Observer \cite{observer} that deal only with complete observations). Among these approaches are ARMS \cite{arms},  SLAF \cite{slaf}, Louga \cite{louga} or Plan-Milner algorithm \cite{planMilner}. ARMS gathers knowledge on the statistical distribution of frequent sets of actions in the plan traces. It then forms a weighted propositional satisfiability problem (weighted SAT) and solves it with a weighted MAX-SAT solver. Unlike ARMS, SLAF is able to learn action models with conditional effects. To that end, SLAF relies on the building of logical constraint formula based on a direct acyclic graph representation. Then, Louga takes also as input plan traces and work with partial noiseless observations. However, Louga is able to learn action models with static properties and negative preconditions. Louga uses a genetic algorithm to learn action effects and an ad-hoc algorithm to learn action preconditions. Then, Plan-Milner uses a classification algorithm based on inductive rule learning techniques: it learns action models with discrete numerical values from partial and noisy observations. Finally, the LOCM family of action model learning approaches \cite{locm,locm2,lop,nlocm} works without information about initial, intermediate and final states. These algorithms extract, from plan traces, parameterized automata representing the behaviour of each object. Then preconditions and effects are generated from these automata.

The last group of works takes as input a set of action sequences randomly generated. Random walk approaches like IRALe \cite{irale} deal with complete but noisy observations. IRALe is based on an online active algorithm to explore and to learn incrementally the action model with noisy observations. Others approaches such as LSO-NIO \cite{lsonio} are able to deal with both partial and noisy observations. LSO-NIO uses a classifier based on a kernel trick method to learn action models. It consists of two steps: (1) it learns a state transition function as a set of classifiers, and (2) it derives the action model from the parameters of the classifiers.

AMLSI differs from the state-of-the-art algorithms in several points. Firstly, AMLSI is one of the few algorithms able to deal with noisy and partial observations. Secondly, AMLSI works with both feasible and infeasible action sequences while most approaches use only feasible action sequences or plan traces. To our best of knowledge, only IRALe and LSO-NIO use failures in action sequences, but AMLSI differentiates feasible and infeasible action sequences. In addition, AMLSI uses random walks to generate its training datasets of both feasible and infeasible action sequences, whereas most of the algorithms like ARMS, Louga and Plan-Milner either only use plan traces or random walks generating feasible action sequences. Thirdly, in terms of expressivity, AMLSI learns PDDL domains including static relations in preconditions as well as negative preconditions. To our best knowledge, only Louga has the same expressivity but it cannot work with noisy traces. Also, AMLSI is an interactive approach. As far as we know,, the only other interactive approach is OpMaker, however these two approaches differ in several ways. Indeed OpMaker takes a partial domain and plan traces as input while AMLSI takes random walks. Moreover, OpMaker only uses feasible actions. Also, the interactive aspect is different, for OpMaker the interactions allow to know the intermediate states, while the interactions allow AMLSI to know if an action is feasible or not for a given state. More importantly, AMLSI is the only algorithm able to learn planning domains accurate enough to be used by planners to solve new planning problems (i.e. that are not in the training sets) with such a level of noise in observations (see Section \ref{sec:exp}).

%% file: parts/problem.tex
\section{Problem Statement}
\label{sec:problem}

We work in the context of classical STRIPS planning \cite{strips}. World states $s$ are modeled as sets of propositions and actions change the world states. Formally, let $S$ be a set of all the propositions modeling properties of world, and $A$ the set of all the possible actions in this world. A state $s$ is a subset of $S$ and each action $a \in A$ is a tuple $(\eta_a, \rho_a, \epsilon^{+}_a, \epsilon^{-}_a)$, where $\eta_a$ is the name of $a$, $\rho_a, \epsilon^{+}_a, \epsilon^{-}_a \subseteq S$ are sets of propositions, and $\epsilon^{+}_a \cap \epsilon^{-}_a = \emptyset$. $\rho_a$ are the preconditions of  $a$, that is, the propositions that must be in the state before the execution of $a$. $\epsilon^{+}_a$ and $\epsilon^{-}_a$ are respectively the positive and the negative effects of  $a$, that is, the propositions that must be added or deleted in $s$ after the execution of the action $a$. An operator is a lifted action described with PDDL.

Let $\gamma : S \times A \rightarrow S$ be the state transition function of a system such that $s' = \gamma(s, a) = (s \cup \epsilon^{+}_a) \setminus \epsilon^{-}_a$. $\gamma(s,a)$ is defined if and only if $\rho_a \subseteq s$. Let $\pi = [a_0, a_1, \dots, a_n]$ be a sequence of actions, and $+$ the concatenation of two sequences. $\Gamma(s_0,\pi)$ is defined recursively as follows:
$$
\resizebox{0.91\hsize}{!}{%
$\Gamma(s_0,\pi) =
\begin{cases}
[s_0]& \text{if } \pi = \emptyset\\
[s_0]& \text{if } \rho_{a_0}\nsubseteq s_0\\
[s_0]+\Gamma(\gamma(s_0, a_0), [a_1,\ldots, a_n])& \text{otherwise}
\end{cases}$
}
$$
$\pi$ is \textit{feasible} given a state $s_0$ if and only if $\Gamma(s_0, [a_0, a_1, \dots, a_n]) = [s_0, s_1, \dots, s_n]$, and $\pi$ and $[s_0, s_1, \dots, s_n]$ have the same length. Otherwise $\pi$ is an {\it infeasible} sequence of actions.  In this paper, we assume that:
\begin{itemize}
\item for all $a \in A$, $(\eta_a, \rho_a, \epsilon^{+}_a, \epsilon^{-}_a)$, the name of $a$ is known but not $\rho_a$, $\epsilon^{+}_a$ and $\epsilon^{-}_a$;
\item $\gamma$ is the state transition function to learn and to express as a set of PDDL lifted operators usable by a planner ;
\item the observations $\Gamma(s_0, \pi)$ are possibly partial and noisy. A partial observation is a state where some propositions are missing, i.e. could either be true or false. A noisy observation is state where the truth value of some propositions is incorrect, i.e. some propositions observed as false should have been observed as true and vice versa.
\end{itemize}

%% file: parts/domain.tex
\section{The AMLSI approach}\label{sec:amlsi}
% \begin{algorithm}[!t]
%   \SetKwInOut{Input}{input }
%   \SetKwInOut{Output}{output }
%   \Input{$A$: Set of actions, $P$: Set of propositions}
%   \Output{$D$: A PDDL domain}
%   $I_{+}, I_{-} \leftarrow generation(A, P);$\\
%   $I^P_{+}, I^P_{-} \leftarrow preprocessing(I_{+}, I_{-}, A);$\\
%   $Aut \leftarrow RPNI(I^P_{+}, I^P_{-});$\\
%   $\mu_A, \mu_P \leftarrow mapping(Aut, I_{+});$\\
%   $D \leftarrow operatorGeneration(Aut, \mu_A, \mu_P);$\\
%   \Repeat{$D = D^{'}$}{
%     $D^{'} \leftarrow D;$\\
%     \Repeat{$D = D^{''}$}{
%         $D^{''} \leftarrow D;$\\
%         $D \leftarrow effectRefinement(D, Aut, \mu_A, \mu_P);$\\
%         $D \leftarrow preconditionRefinement(D);$\\
%     }
%     $D \leftarrow tabuSearch(I_{+}, I_{-}, D);$\\
%   }
%   \Return{$D$}
%   \caption{\label{algo:amlsi}AMLSI}
% \end{algorithm}
The AMLSI algorithm takes as input the set of action $A$ and the set  of  propositions $S$, builds two datasets by interactions with a state machine and returns a PDDL domain. After building datasets, the AMLSI approach performs the learning phase which is composed of three steps. We begin by a grammar induction step. Then we generate, from the grammar previously learned, PDDL operators. Finally, we refine the domain.

\subsection{Dataset generation}\label{sec:generation}
The objective of this step is to build two training datasets: $I_{+}$ (positive samples) containing the feasible action sequences and the corresponding observations, and $I_{-}$ (negative samples) containing the infeasible action sequences: at a given state $s$, we query the system about the feasibility of an action $a$ randomly chosen in $A$. If $a$ is feasible, the current state is observed and we add $a$ to the current $\pi$. This random walk is iterated until $\pi$ reaches an arbitrary length, and added to $I_{+}$. If $a$ is infeasible in the current state, the concatenation of $\pi$ and $a$ is added to $I_{-}$.

\subsection{Grammar Induction}\label{subsec:domain:grammar}

In the second step, we use the RPNI algorithm \cite{rpni} to learn the regular grammar based on $I_{+}$ and $I_{-}$ inputs. RPNI has a polynomial complexity and is optimal: it returns the smallest automaton (a regular grammar can be represented as a deterministic finite automaton) accepting all the positive samples $I_{+}$ and rejecting all the negative samples $I{-}$ when $I_{+}$ and $I{-}$ are "characteristic" \cite{rpni}. Formally, the deterministic finite automaton learned is a quintuple $<A, N, n_0, \gamma, F>$, where $A$ is the set of actions, $N$ is the set of nodes, $n_0 \in N$ is the initial node, $\gamma$ is the node transition function, and $F \subseteq N$ is the set of final nodes.

Finally, to prepare the grammar induction, we built $I_{+}^P$ and $I_{-}^P$, which are samples $I_{+}$ and $I_{-}$ extended tanks a preprocessing steps. Pairwise constraints (PC) are pairs of action that cannot be consecutive in a sequence. These constraints are based on the fact that for an action to be feasible a certain number of resources must be produced (add list) and others must be consumed (del list). For instance, in the gripper domain, $move(r_1~r_2)$ is never followed by $pick(b~r_1~grip)$ because $(at-robby~r_1)$ must be true to execute $pick(b~r_1~grip)$, and after the execution of $move(r_1~r_2)$, $(at-robby~r_1)$ is always false, therefore $pick(b~r_1~grip)$ cannot follow $move(r_1~r_2)$. PC computation is based on $I_{+}$: we assume that only pairs of actions in $I_{+}$ are possible:
\begin{equation*}
\scalebox{0.8}[1]{$
\begin{array}{l}
\{\forall a_i,a_j \in A^2, a_i,a_j \in I_{-}^p~iff~\nexists \pi \in I_{+}~s.t.~\pi=[\pi_1,a_i,a_j,\pi_2]\}\\
\end{array}
$}
\end{equation*}

%\subsubsection{Sequence prefixes.} Any prefix of a feasible sequence is itself feasible. The correct grammar has to accept them, so prefixes are added to $I_{+}^P$. For instance, if $[move(r_1~r_2), move(r_2~r_1), pick(b~r_1~left)] \in I_{+}$, $[move(r_1~r_2)$, $move(r_2~r_1)$, $pick(b~r_1~left)]$, $[move(r_1~r_2),move(r_2~r_1)]$ and $[move(r_1~r_2)]$ are in  $I_{+}^P$.

\subsection{Operator generation}\label{sec:op_induc}
Operator generation is based on four steps:
\subsubsection{Observation mapping}
Once the automaton is induced, we need to know which node of the automaton corresponds to which observed state. To do that, we input in the automaton all the positive samples in $I_{+}$ and map the pairs "node, action" in the automaton with the pairs "state, action" in $I_{+}$. There are two different mappings: the mapping (A)nte $\mu_A$ and the mapping (P)ost $\mu_P$. $\mu_A(n,a)$ (resp. $\mu_P(n,a)$) gives the state before (resp. after) the execution of the transition $a$ in node $n$. Then, we compute a reduced mapping $\bigcap \mu_A$ (resp $\bigcap \mu_P$), which contains the common propositions of all the states for a given $(n,a)$. For instance, consider the classical gripper planning domain: $(at~b~r_1)$ is in $\bigcap \mu_A(0,pick(b~r_1~grip))$ iff $(at~b~r_1)$ is in all $\mu_A(0, pick(b~r_1~grip))$. Likewise, we remove from $\bigcap \mu_A(0,pick(b~r_1~grip))$ and $\bigcap \mu_P(0,pick(b~r_1~grip))$ all the propositions whose parameters are not a subset of $(b~r_1~grip)$, the parameters of $pick(b~r_1~grip)$ (classical planning assumption \cite{strips}).

%The advantage to use the common pattern in the mapping is that it reduces the noise in the mapping. Indeed, when we have a high level of compression, it is unlikely that a proposition present in the mapping is a misjudged proposition.
\subsubsection{Precondition generation}
%The algorithm \ref{algo:prec} describes how we generate operator preconditions. To learn the preconditions of an operator $o$, we compute the reduced mapping of all the nodes where an action $a$, instantiating $o$, is feasible. At line 8, $\rho_a$ is the reduced mapping of all the $\mu_A$ mappings of the action $a$. The generalization of $\rho_a$ at line 9 replaces all the constant symbols by variables (for instance, $(b~r_1~grip)$ by $(?obj~?room~?gripper)$).
%
%\begin{algorithm}[!t]
%  \SetKwInOut{Input}{input }
%  \SetKwInOut{Output}{output }
%  \Input{$o$: an operator}
%  \Output{$\rho_o$: preconditions of $o$}
%  $\Xi \leftarrow \{\};$\\
%  \For{$a$ instance of $o$}{
%    $\tau \leftarrow \{\}$;\\
%    \For{$n$ in the automaton}{
%      \If{$a$ is an outgoing edge of $n$} {
%        \If{$\bigcap\mu_A(n,a) \neq \emptyset$}{
%          $\tau \leftarrow \tau \cup \{\bigcap\mu_A(n,a)\};$\\
%        }
%      }
%    }
%    $\rho_a \leftarrow \bigcap(\tau);$\\
%    $\Xi \leftarrow \Xi \cup \{generalize(\rho_a)\};$\\
%  }
%  $\rho_o \leftarrow \bigcap(\Xi);$\\
%  \Return{$\rho_o$}
%  \caption{\label{algo:prec}Precondition generation}
%\end{algorithm}
To learn the preconditions of an operator $o$, we find all propositions that are always present in the reduced mapping of all the nodes where an action $a$, instantiating an operator $o$, is feasible. Formally,
$p~(resp~\neg p) \in \rho_o$ if and only if $\forall a$ instance of $o$ :
\begin{equation*}
\forall (n,a): p~(resp~\neg p) \in \bigcap\mu_A(n,a)
\end{equation*}
\subsubsection{Effect generation}
To learn the effects of an operator $o$, we find all propositions that are always (resp never) present before the execution of an action $a$,  instantiating an operator $o$, in the automaton and never (resp always) present after the execution. Formally,
$\neg p~(resp~p) \in \epsilon_o^{-}~(resp~\epsilon_o^{+})$ if and only if $\forall a$ instance of $o$ :
\begin{equation*}
\begin{array}{ll}
\forall (n,a,n'): &p~(resp~\neg p) \in \bigcap\mu_A(n,a)\\
                  &\wedge \neg p~(resp~p) \in \bigcap\mu_P(n',a)\\
\end{array}
\end{equation*}

\subsection{Operator refinement}
\label{sec:refinement}
The refinement steps of the PDDL operators are necessary because the observations are partial and noisy. The refinement is divided into 3 substeps:

\subsubsection{Effect refinement}
This step ensures that the generated operators allow to regenerate the induced grammar. We use the reduced mappings $\bigcap \mu_A$ to verify that for each couple of consecutive actions $a$ and $a'$, the effects of the action $a$ generate the preconditions of action $a'$. If it is not the case, we add in the effects of $a$ the propositions satisfying the preconditions of $a'$. For instance, suppose we have $n' = \gamma(n,move(r_1~r_2))$ and $n'' = \gamma(n',pick(b~r_2~grip))$. Now suppose we have $\neg (at-robby~r_2) \in \bigcap \mu_A(n,move(r_1~r_2))$, $(at-robby~r_2) \in \rho_{pick(b~r_2~grip)}$ and $(at-robby~r_2) \not\in \epsilon^{+}_{move(r_1~r_2)}$. We need to have $(at-robby~r_2) \in \epsilon^{+}_{move(r_1~r_2)}$ in order to make $\gamma(n,move(r_1~r_2))$ and $\gamma(n',pick(b~r_2~grip))$ feasible. Thus, we add $(at-robby~?to)$ to $\epsilon^{+}_{move(?from~?to)}$.

\subsubsection{Precondition refinement}
In this step, we assume like \cite{arms} that the propositions of the negative effects must be in the action preconditions. Thus for each negative effect in an operator, we add the
corresponding proposition in the preconditions. For instance, suppose $\neg (at-robby~?from) \in \epsilon^{-}_{move(?from~?to)}$, then $(at-robby~?from) \in \rho_{move(?from~?to)}$ after
refinement.

Since effect refinement depends on the preconditions and precondition refinement depends on the effects, we repeat these two steps until convergence. They converge because the adding of preconditions is limited by the effects, and the adding of effects is limited by the preconditions of the next action in the induced automaton (In our experiments, see Section \ref{sec:exp}, less than 10 iterations are needed to converge).

\subsubsection{Tabu search}
Finally, we perform a Tabu Search to improve the PDDL operators independently of the induced grammar, on which operator generation is based.

The neighborhood of a candidate domain is the set of domains where a precondition or an effect is added or removed. And the search space of the tabu search is the set of all possible domains compatible with the PDDL syntax constraints \cite{arms}. The fitness function used to evaluate a candidate set $D$ of PDDL operators is:
\begin{equation*}
\scalebox{0.95}[1]{$
\begin{array}{ll}
J(D | I_{+}, I_{-}) = & J_\rho(D | I_{+}) + J_\epsilon(D | I_{+}) +\\
&J^{+}(D | I_{-}) + J^{-}(D | I_{-})\\
\end{array}
$}
\end{equation*}
where :
\begin{itemize}
\item \scalebox{0.9}[1]{$J_\rho(D | I_{+}) = \sum\limits_{\pi \in I_{+}}
\sum\limits_{s \in \Gamma(s_0, \pi)} Accept(\rho_a, s) - Reject(\rho_a, s)$} computes the
fitness score for the preconditions of the actions $a$. $Accept(\rho_a, s)$ counts the number of
positive and negative preconditions in the observed state $s$, and
$Reject(\rho_a, s)$ counts the number of positive and negative preconditions that are not in $s$.
\item \scalebox{0.9}[1]{$J_\epsilon(D | I_{+}) = \sum\limits_{\pi \in I_{+}}
\sum\limits_{s \in \Gamma(s_0, \pi)} Equal(s, \hat{s}) - Different(s, \hat{s})$} computes the
fitness score for the effects of the actions $a$. $s$ are the observed states and $\hat{s}$ are the states obtained by applying the actions of $\pi$. $Equal(s, \hat{s})$ counts the number of similar propositions in $s$ and $\hat{s}$, and $Different(s, \hat{s})$ counts the differences.
\item \scalebox{0.85}[1]{$J^{+}(D | I_{+}) = \sum\limits_{\pi \in I_{+}}
|\pi| \times \mathbbm{1}_{Accept(D, \pi)}$ where $\mathbbm{1}_{Accept(D, \pi)} = 1$}
if and only if $D$ can generate the positive sample $\pi$. $|\pi|$ is the length of $\pi$. $J^{+}(D | I_{+})$ is weighted by the length of $\pi \in I_{+}$ because $I_{+}$ is smaller than $I_{-}$
\item \scalebox{0.95}[1]{$J^{-}(D | I_{-}) = \sum\limits_{\pi \in I_{-}}
\mathbbm{1}_{Accept(D, \pi_{+}) \wedge Reject(D, \pi)}$}. As detailed in section \ref{sec:generation}, the negative sample $\pi \in I_{-}$ is a sequence of $n+1$ actions where the $n$ first actions are a prefix of a sequence in $I_{+}$: $\pi_{+}$ is this prefix. $\mathbbm{1}_{Accept(D, \pi_{+}) \wedge Reject(D, \pi)} = 1$ if and only if $D$ can generate $\pi_{+}$ and not $\pi$.
\end{itemize}

%Fitness functions $J_\rho(D | I_{+})$ and $J_\epsilon(D | I_{+})$ measure the ability of $D$ to explain the observed states whereas fitness functions $J^{+}(D | I_{+})$ and $J^{-}(D | I_{-})$ measure its ability to regenerate the induced grammar.

Once the Ttabu search is done, we repeat all the refinement steps until convergence.

%% file: parts/experiment.tex
\section{Experiments}\label{sec:exp}
\input{parts/benchmark.tex}
\input{parts/automaton.tex}
\input{parts/amlsi_lsonio.tex}
Our experiments are based on 7 IPC domains: Blocksworld, Gripper, Peg Solitaire, Parking, Zenotravel, Sokoban and Neg-Elevator. Neg-Elevator is a modified version of Elevator with negative preconditions to show AMLSI ability to learn them. All the used benchmarks are STRIPS domain. Table \ref{tab:domain} shows our experimental setup.

We deliberately chose the size of the test sets larger than the learning sets to show AMLSI's ability to learn accurate domains with small datasets. The training and test sets are generated as explained in Section \ref{sec:generation}. In the training sets, we generate positive action sequences with a length randomly chosen between $10$ and $20$, and in the test sets, we generate positive action sequences with a length randomly chosen between $1$ and $100$. $I_{-}$ are bigger than $I_{+}$ because it is more likely to generate infeasible actions.

We test each IPC domain with three different initial states over five runs, and we used five seeds randomly generated for each run. Tabu search is computed over 200 runs. For each IPC domain, we generate observed states by randomly removing a fraction of the propositions (partial states) and by randomly modifying their truth values. All tests were performed on an Ubuntu 14.04 server with a multi-core Intel Xeon CPU E5-2630 clocked at $2.30$ GHz with 16GB of memory.

\subsection{Evaluation Metrics}
\label{subsec:Metrics}
AMLSI is evaluated with four metrics of the literature:

\subsubsection{Syntactical error} The syntactical error $error(o)$ \cite{zhuo10} for an operator $o$ is defined as the number of extra or missing predicates in the preconditions\footnote{Note that we compute the syntactical error without taking into account the negative preconditions of the operators because some of the chosen IPC domains do not have them.} $\rho_o$, the positive effects $\epsilon_o^{+}$ and the negative effects $\epsilon_o^{-}$ divided by the total number of possible predicates. The syntactical error for a domain with a set of operator $O$ is: $E_{\sigma} = \frac{1}{|O|} \sum_{o \in O} error(o)$.
%\footnote{Note that we compute the syntactical error without taking into account the negative preconditions of the operators because some of the chosen IPC domains do not have them.}
\subsubsection{Precondition error rate} The precondition error rate \cite{arms} computes the rate of preconditions that are not satisfied in the positive test set. This metric measures the quality of the preconditions in the learned domain. It is computed as follows:
$$E_{\rho} = \sum\limits_{\pi \in E^{+}}
\frac{\sum\limits_{a \in \pi} error(\rho_a, \hat{s})}
{\sum\limits_{a \in \pi}|\rho_a|}
$$
$error(\rho_a, \hat{s}) =$ $| \{p \in \rho_a \wedge \neg p \in \hat{s}\}| + | \{\neg p \in \rho_a \wedge p \in \hat{s}\}|$ gives the number of positive and negative preconditions $p$ in actions $a$ that are not satisfied in the observed state $\hat{s}$.

\subsubsection{Effect error rate} The effect error rate \cite{louga} computes the error rate on the effects of the learned domain. It is computed as follows:
$$E_{\epsilon} = \sum\limits_{\pi \in E^{+}}
\frac{\sum\limits_{a \in \pi} error(\epsilon_a, \hat{s})}
{\sum\limits_{a \in \pi}|\epsilon_a|}$$
$error(\epsilon_a, \hat{s}) =$ $| \{p \in \epsilon^{+}_a \wedge \neg p \in
\hat{s}\}| + | \{\neg p \in \epsilon^{-}_a \wedge p \in \hat{s}\}|$ gives the number of
positive and negative effects $p$ in the actions $a$ that are not
satisfied in the observed state $\hat{s}$.

\subsubsection{Accuracy} The accuracy \cite{rim} quantifies the usability of the learned model for planning. Most of the works addressing the problem of learning planning domains uses the syntactical error to quantify the performance of the learning algorithm. However, domains are learned to be used for planning, and it often happens that one missing precondition or effect makes them unable to solve new planning problems. Formally, the accuracy $Acc = \frac{N}{N^{*}}$ is the ratio between $N$ the number of correctly solved problems with the learned domain and $N^{*}$ the total number of problems to solve. The accuracy is computed over 20 problems. Although rarely used, we argue that the accuracy is the most important metric because it measures to what extent a learned domain is useful in practise for planning. Problem are solved with Fast Downward 19.06 \cite{fastDown}. Plan validation is realized with the automatic validation tool used in the IPC competition: VAL \cite{val}. Finally, we also report in our results the ratio of solved problem that are not necessarily correctly solved, i.e. the ratio of problems where the learned domain was able to find a plan even if the found plan was not validated by the ground truth domain.

\subsection{Results}
\label{sec:impact-observations}

In order to study the performances of AMLSI with respect to noisy and partial observations, we use four different experimental scenarios:
\begin{enumerate}
\item Complete intermediate observations (100\%) and no noise (0\%), see Table \-- \ref{tab:res_complete}.
\item Complete intermediate observations (100\%) and high level of noise (20\%), see Table \-- \ref{tab:res_complete}.
\item Partial intermediate observations (25\%) and no noise (0\%), see Table \-- \ref{tab:res_partial}.
\item Partial intermediate observations (25\%) and high level of noise (20\%), see Table \-- \ref{tab:res_partial}.
\end{enumerate}

\subsubsection{Impact of noisy and partial observations}

The results of the first scenario (complete intermediate observations and no noise) show that AMLSI perfectly learns the preconditions and the effects of the operators of the IPC domains. Note that for 5 IPC domains, Peg-Solitaire, Parking and Sokoban, the syntactical error is not equal to $0$. This is because AMLSI learns preconditions that are not in the IPC domain. On the other hand, the obtained accuracy is optimal for all domains. This means that the domains learned with AMLSI can be used to solve all the problems of the IPC domains.

The results of the second scenario (complete intermediate observations and high level of noise (20\%) are almost similar to the first scenario. Noise slightly reduces the quality of learning in terms of syntactical error and accuracy for three IPC domains (Blocksworld, Peg Solitaire and Zenotravel). The impact of noise on the performance of AMSLI with complete intermediate observations is low.

The results of the third scenario (partial intermediate observations (25\%) and no noise) are almost similar to the first scenario. Partial observation slightly reduces the quality of learning in terms of syntactical error and accuracy for one IPC domain (Peg Solitaire). The impact of partial observation on the performance of AMLSI with noiseless observations is low.

Finally, the results of the fourth scenario (partial intermediate observations (25\%) and high level of noise (20\%)) show that partial and noisy intermediate observations downgrade the global performances of AMLSI. However, they remain high for all the IPC domains and metrics, and specifically for the accuracy. Despite partial and noisy observations, AMLSI is able to generate accurate PDDL domains.

Moreover, we can observe that the domain including both positive and negative preconditions (Neg-Elevator) is easier to learn that domains including only positive preconditions. This is due to the fact that negative preconditions have a stronger impact on the fitness score of the Tabu search. Then, for domain using only positive preconditions, the easiest domain to learn (Gripper) is a domain with the highest level of compression (see Table \-- \ref{tab:automata}) in the induced automaton, and the most difficult domain to learn (Peg Solitaire) is one of the domain with the lowest level of compression.

% \subsubsection{Impact of the initial state}
% \label{sec:impact-initial-state}
%
% Table \-- \ref{tab:std} shows the standard deviation between the different initial states for the fourth experimental scenario and for each performance metrics. We can observe that for some domains the choice of the initial state has a significant impact.
%
% The performance gap between initial states can be explained by the different complexities of the grammars to induce. For instance, for the Peg Solitaire domain, the automaton learned with the first initial state contains generally 20 nodes and 40 transitions and the compression level varies between 7 and 9, while the automaton learned with the third initial state contains generally 50 nodes and 70 transitions and the compression level varies between 4 and 6. It is, therefore, easier to learn the domain with the first initial state than the third initial states.

\subsubsection{Ablation study}
In addition, we perform an ablation study of the AMLSI approach (see Tables \ref{tab:res_complete} and \ref{tab:res_partial}). We test each part of the AMLSI approach independently:
\begin{enumerate}
\item \textbf{Generation Step:} We learn domains by taking into account only the operator induction step. We can observe that for the majority of domains, this step is sufficient for the first experimental scenario (complete and noiseless observations). However when observations are partial and/or noisy, this step is not able to learn domains.

\item \textbf{Simple refinement:} We learn refined domain by taking into account only preconditions/effects precondition steps, i.e. without the Tabu search. As for the generation step, this refinement is generally able to learn domains when observations are complete and noiseless. In addition, this refinement is able to learn some domains (Blocksworld and Gripper) when observations are partial and noiseless. However when observations are noisy, this refinement is not able to learn domains. This is due to the fact that, during the mapping, noisy propositions will delete a large amount of information. The effect refinement steps will therefore no longer be able to detect all of the missing effects. In addition, when observations are noisy, there is a high risk that this step will detect additional effects.

\item \textbf{Tabu search alone:} We learn domains with only the Tabu search, i.e. without the operator induction and preconditions/effects refinement steps. We can observe that for the majority of domains, Tabu search alone is not able to learn domains, whatever the experimental scenario. We can also note that the Tabu search alone generally learns the best domains when observations are noisy. Finally we can note that it is the combination of the Tabu search and the effects and preconditions refinement steps that allow AMLSI to learn accurate domains in each experimental scenario.
\end{enumerate}

\subsubsection{Grammar induction without pairwise constraints}

Then, we test a variant of the AMLSI algorithm where the grammar induction was performed without pairwise constraints (noted \textit{Without PC} in Tables \ref{tab:res_complete} and \ref{tab:res_partial}). First of all, we can observe that the results are generally deteriorated for each domain and for each experimental scenario.

The performance gap can be explained by the quality of induced grammar. Indeed, as we have seen in section \ref{sec:amlsi}, RPNI is optimal if and only if samples are "characteristic". The construction of a characteristic sample is not feasible a priori and we can not assume that the dataset generation produces characteristic samples. That implies that grammars induced without PC are more general that grammars induced with PC, and there are therefore more unfeasible sequences, i.e. sequences present in the grammar induced and not present in the ground truth grammar, in grammars induced without PC. These unfeasible sequences causes noises in the effects refinement, i.e. extra effects are detected during the effects refinement step.

Finally, we can note that, when grammars are induced without PC, domains are generally better when observations are noisy. This due to the fact that when observations are noisy, effects refinement step detected less extra effects.

\subsubsection{Comparison with LSO-NIO}
LSO-NIO has been tested with $I_{+}$ and $I_{-}$. Random walks including action failures are obtained by merging $I_{+}$ and $I_{-}$. Note that, in our experiment LSO-NIO's training set contains between $2000$ and $5000$ actions with a majority of failed actions, while LSO-NIO was tested with a training set containing $20000$ actions with an equal mixture of successful and unsuccessful actions. Also, for our experimentation, there are fewer objects in initial states than for the experimentation of \cite{lsonio}.

We can observe in Tables \ref{tab:res_complete} and \ref{tab:res_partial} that AMLSI outperforms LSO-NIO. These results can be explained in several ways. First of all, we generally have a lot of negative information than positive information. This is beneficial for AMLSI because it makes it possible to have a good automaton, and it makes it possible to lead efficiently the Tabu search. This bias LSO-NIO because the updated weights of the different classifiers need more positive information. In addition, LSO-NIO learns effects and preconditions by taking into account actions one by one. While AMLSI, during the Tabu search, learns effects and preconditions by taking into account all action sequences. Finally, we observe that LSO-NIO does not learn static relations in precondition with our samples.

%% file: parts/benchmark.tex
\begin{table*}[!t]
\begin{center}
\resizebox{\textwidth}{!}{
\begin{tabular}{|c|c|c|c|c|c|c|c|c|c|c|c|c|c|}
\hline
Domain & $\#Operators$ & $\#Predicates$ &$\#Objects$ & $\#Actions$ & $\#Propositions$& $|I_{+}|$ & $|I_{-}|$ & $|\pi_{+}|$ & $|\pi_{+}|$ & $|E_{+}|$ & $|E_{-}|$ & $|e_{+}|$ & $|e_{+}|$\\ \hline
Blocksworld &$ 4 $&$ 5 $& $3$ &$18$&$16$&$ 30 $&$ 2421.3 $&$ 15 $&$ 8.3 $&$ 100 $&$ 26209.5 $&$ 49 $&$ 33.6$\\ \hline
Gripper &$ 3 $&$ 4 $& $5$&$10$&$8$&$ 30 $&$ 1168.1 $&$ 15.2 $&$ 8.3 $&$ 100 $&$ 12940.3 $&$ 50.7 $&$ 33.7$\\ \hline
%Hanoi &$ 4 $&$ 7 $& $6$ &$24$&$25$&$ 30 $&$ 3011.3 $&$ 14.8 $&$ 8.2 $&$ 100 $&$ 34780.7 $&$ 50.6 $&$ 33.8$\\ \hline
%N-Puzzle &$ 1 $&$ 3 $& $7$ &$24$&$24$& $ 30 $&$ 3354.1 $&$ 15.2 $&$ 8.4 $&$ 100 $&$ 36613.4 $&$ 49.9 $&$ 34$\\ \hline
Peg Solitaire &$ 3 $&$ 5 $& $9$ &$38$&$45$&$ 30 $&$ 4486.8 $&$ 7.1 $&$ 5.4 $&$ 100 $&$ 14509.5 $&$ 6.9 $&$ 5.3$\\ \hline
Parking &$ 4 $&$ 5 $& $6$ &$60$&$24$&$ 30 $&$ 5729.53 $&$ 15 $&$ 8.5 $&$ 100 $&$ 65216.6 $&$ 50.6 $&$ 34$\\ \hline
Zenotravel &$ 5 $&$ 5 $& $7$ &$14$&$10$& $ 30 $&$ 1631.4 $&$ 15.1 $&$ 8.4 $&$ 100 $&$ 17850.3 $&$ 49.6547 $&$ 33.9$\\ \hline
Sokoban &$ 2 $&$ 4 $&$14$&$36$&$51$&$ 30 $&$ 4634.33 $&$ 15 $&$ 8.2 $&$ 100 $&$ 51832.7 $&$ 51.3 $&$ 33.4$\\ \hline
%Neg Visit All &$ 2 $&$ 3 $& $4$ &$9$&$13$&$ 30 $&$ 1275.7 $&$ 15 $&$ 8.9 $&$ 100 $&$ 16666.1 $&$ 51 $&$ 35.6$\\ \hline
Neg Elevator &$ 4 $&$ 6 $& $5$ &$8$&$13$&$ 30 $&$ 1050.7 $&$ 15.1 $&$ 8.6 $&$ 100 $&$ 13086.6 $&$ 51 $&$ 35.7$\\ \hline
\end{tabular}
}
\caption{\label{tab:domain} Benchmark domain characteristics (from left to right): number of operators, number of predicates, number of objects in each initial states, number of actions in each initial states, number of propositions in each initial states, average size of $|I_{+}|$ and $|I_{-}|$ training sets, the average length of the positive (resp. negative) training sequences $\pi_{+} \in I_{+}$ (resp. $\pi_{-} \in I_{-}$), average size of $|E_{+}|$ and $|E_{-}|$ test sets, the average length of the positive (resp. negative) test sequences $e_{+} \in E_{+}$ (resp. $e_{-} \in E_{-}$).}
\end{center}
\end{table*}

%% file: parts/automaton.tex
\begin{table}[!t]
\begin{center}
\resizebox{0.45\textwidth}{!}{
\begin{tabular}{|c|c|c|c|c|}
\hline
Domain        & $\#States$ & $\#Nodes$ & $\#Transitions$ & Compression level \\ \hline
Blocksworld   & $449.7$    & $23$      & $43.4$          & $19.7$            \\ \hline
Gripper       & $455$      & $8$       & $16$            & $58.9$            \\ \hline
%Hanoi         & $443.6$    & $27.9$    & $51.47$         & $16.7$            \\ \hline
%N-Puzzle      & $455.4$    & $12.7$    & $24.7$          & $35.9$            \\ \hline
Peg-Solitaire & $212$    & $34.6$    & $46.7$          & $8.1$             \\ \hline
Parking       & $450.4$    & $78.7$    & $175.3$         & $6.1$             \\ \hline
Zenotravel    & $453.1$    & $24.1$    & $53$            & $19.5$            \\ \hline
Sokoban       & $450.7$    & $31.3$    & $64.1$          & $16.8$            \\ \hline
%Neg Visit All & $449.7$    & $39.7$    & $64.5$          & $12.4$            \\ \hline
Neg Elevator  & $451.5$    & $24.5$    & $44.4$          & $18.9$            \\ \hline
\end{tabular}
}
\caption{\label{tab:automata} Induced automaton characteristics (from left to right): average number of states in the learning set, average number of nodes, average number of transitions, compression level, i.e. average number of states per node.}
\end{center}
\end{table}

%% file: parts/amlsi_lsonio.tex
\begin{table*}[!t]
\begin{subtable}{0.9\textwidth}
\begin{center}
\resizebox{0.7\textwidth}{!}{
\begin{tabular}{|c|c||c|c|c|c|c||c|c|c|c|c|}
\hline
\multicolumn{2}{|c||}{Noise} & \multicolumn{5}{|c||}{$0\%$} & \multicolumn{5}{|c||}{$20\%$} \\ \hline
Domain & Algorithm & $E_\rho (\%)$ & $E_\epsilon (\%)$ & $E_\sigma (\%)$ & $Solved (\%)$ & $Acc (\%)$ & $E_\rho (\%)$ & $E_\epsilon (\%)$ & $E_\sigma (\%)$ & $Solved (\%)$ & $Acc (\%)$  \\ \hline\hline
\multirow{5}{*}{Blocksworld}
& AMLSI  & $0$ & $0$ & $0$ & $100$ & $100$
         & $0.8$ & $0.8$ & $0.6$ & $93.7$ & $93.7$\\
& Generation step
         & $0$ & $0$ & $0$ & $100$ & $100$
         & $0$ & $0$ & $33.25$ & $0$ & $0$\\
& Simple refinement
         & $0$ & $0$ & $0$ & $100$ & $100$
         & $7.7$ & $6.6$ & $26.3$ & $20$ & $0$\\
& Tabu search alone
         & $0$ & $0$ & $9.8$ & $86.7$ & $13$
         & $0.3$ & $0.5$ & $9.6$ & $87$ & $13$\\
& Without PC
          & $19$ & $29$ & $22$ & $27.7$ & $27.7$
          & $12.3$ & $21.3$ & $18.5$ & $53.3$ & $34.3$\\
& LSO-NIO & $0$ & $0$ & $0$ & $100$ & $100$
         & $13.5$ & $14.3$ & $20.1$ & $0.3$ & $0$\\ \hline \hline
\multirow{6}{*}{Gripper}
& AMLSI  & $0$ & $0$ & $0$ & $100$ & $100$
         & $0$ & $0$ & $0$ & $100$ & $100$\\
& Generation step
         & $0$ & $0$ & $0$ & $100$ & $100$
         & $0$ & $0$ & $46.9$ & $0$ & $0$\\
& Simple refinement
         & $0$ & $0$ & $0$ & $100$ & $100$
         & $0$ & $0$ & $46.9$ & $0$ & $0$\\
& Tabu search alone
         & $0$ & $0$ & $0$ & $100$ & $100$
         & $0$ & $0$ & $0$ & $100$ & $100$\\
& Without PC
         & $0$ & $0$ & $0$ & $100$ & $100$
         & $0$ & $0$ & $0$ & $100$ & $100$\\
& LSO-NIO & $0$ & $0$ & $5.6$ & $100$ & $0$
         & $6$ & $7$ & $22$ & $33.3$ & $0$\\ \hline \hline
% \multirow{6}{*}{Hanoi}
% & AMLSI  & $0$ & $0$ & $0.9$ & $100$ & $100$
%          & $0$ & $0$ & $0.9$ & $100$ & $100$ \\
% & Generation step
%          & $0$ & $0$ & $0.9$ & $100$ & $100$
%          & $0$ & $0$ & $37.8$ & $4.3$ & $0$\\
% & Simple refinement
%          & $0$ & $0$ & $0.9$ & $100$ & $100$
%          & $0$ & $0$ & $34.8$ & $5$ & $0$\\
% & Tabu search alone
%          & $0.8$ & $0$ & $12.7$ & $76.7$ & $0$
%          & $0$ & $0$ & $12.8$ & $71.33$ & $0$\\
% & Without PC
%          & $15.6$ & $30.1$ & $21.8$ & $8.3$ & $6.7$
%          & $0$ & $7.1$ & $14.5$ & $68.7$ & $13.3$\\
% & LSO-NIO & $0$ & $0$ & $9.2$ & $100$ & $0$
%          & $6.2$ & $6.5$ & $21.3$ & $19.7$ & $0$\\ \hline \hline
% \multirow{6}{*}{N-Puzzle}
% & AMLSI  & $0$ & $0$ & $5.6$ & $100$ & $100$
%          & $0$ & $0$ & $5.6$ & $100$ & $100$ \\
% & Generation step
%          & $0$ & $0$ & $5.6$ & $100$ & $100$
%          & $0$ & $0$ & $37.4$ & $13.3$ & $0$\\
% & Simple refinement
%          & $0$ & $0$ & $5.6$ & $100$ & $100$
%          & $0$ & $0$ & $36$ & $26.7$ & $1.33$\\
% & Tabu search alone
%          & $0$ & $0$ & $5.6$ & $100$ & $100$
%          & $0$ & $0$ & $5.6$ & $100$ & $100$\\
% & Without PC
%          & $1.1$ & $1.7$ & $6.7$ & $93.3$ & $93.3$
%          & $0$ & $0$ & $5.6$ & $100$ & $100$\\
% & LSO-NIO & $0$ & $0$ & $5.6$ & $100$ & $20$
%          & $0$ & $0$ & $10.7$ & $88$ & $15.3$\\ \hline \hline
\multirow{6}{*}{Peg Solitaire}
& AMLSI  & ${0}$ & $0$ & ${4.2}$ & $100$ & $100$
         & $2.1$ & ${0.7}$ & $7.4$ & $99.3$ & $96.3$\\
& Generation step
         & ${0}$ & $0$ & $4.2$ & $100$ & $100$
         & ${0}$ & $0$ & $24.3$ & $0$ & $0$\\
& Simple refinement
         & ${0}$ & $0$ & $4.2$ & $100$ & $100$
         & ${1.5}$ & $0$ & $22.5$ & $1$ & $0$\\
& Tabu search alone
         & $2.4$ & $0$ & $9.5$ & $100$ & $100$
         & $4.9$ & $0.1$ & $13.9$ & $93$ & $81.7$\\
& Without PC
         & $9.4$ & $15.4$ & $13.2$ & $65.7$ & $60$
         & $4.4$ & $3.6$ & $10.3$ & $92.3$ & $89.7$\\
& LSO-NIO & $0$ & $0$ & $3.8$ & $100$ & $0$
         & $8.3$ & $10.9$ & $18.8$ & $48$ & $0$\\ \hline \hline
\multirow{6}{*}{Parking}
& AMLSI  & $0$ & $0$ & $3.9$ & $100$ & $100$
         & $0$ & $0$ & $3.9$ & $100$ & $100$\\
& Generation step
         & $0$ & $0$ & $3.9$ & $100$ & $100$
         & $0$ & $0$ & $30.2$ & $4.3$ & $0$\\
& Simple refinement
         & $0$ & $1$ & $4.2$ & $100$ & $77.3$
         & $0$ & $0$ & $30.8$ & $4.3$ & $0$\\
& Tabu search alone
         & $0$ & $0$ & $8.2$ & $85$ & $65$
         & $0$ & $0$ & $8.1$ & $84.7$ & $63.7$\\
& Without PC
         & $29.7$ & $42$ & $41.2$ & $0.7$ & $0.3$
         & $10.4$ & $10$ & $12.5$ & $77$ & $67.33$\\
& LSO-NIO & $0$ & $0$ & $6.5$ & $80$ & $4$
         & $14.9$ & $27.3$ & $25$ & $21.3$ & $0$\\ \hline \hline
\multirow{6}{*}{Zenotravel}
& AMLSI  & $0$ & $0$ & $0$ & $100$ & $100$
         & $0.1$ & $0$ & ${0.2}$ & ${100}$ & $99.3$\\
& Generation step
         & ${0}$ & $0$ & $0$ & $100$ & $100$
         & $0$ & $0$ & $27.4$ & $14.7$ & $0$\\
& Simple refinement
         & ${0}$ & $0$ & $0$ & $100$ & $100$
         & $0.3$ & $1.9$ & $22.4$ & $21.3$ & $0$\\
& Tabu search alone
         & ${0}$ & $0$ & $1.8$ & $100$ & $73.3$
         & ${0}$ & $0$ & $5.4$ & $100$ & $33.3$\\
& Without PC
         & $5$ & $14.5$ & $9.6$ & $69$ & $40$
         & $2.1$ & $9.4$ & $8.4$ & $74.7$ & $66.7$\\
& LSO-NIO & $0$ & $0$ & $9.4$ & $100$ & $0$
         & $10.1$ & $10$ & $21.3$ & $51$ & $0$\\ \hline \hline
\multirow{6}{*}{Sokoban}
& AMLSI  & $0$ & $0$ & $3.9$ & $100$ & $100$
         & $0$ & $0$ & $3.9$ & $100$ & $100$\\
& Generation step
         & $0$ & $0$ & $3.9$ & $100$ & $100$
         & $0$ & $0$ & $20.1$ & $13.3$ & $0$\\
& Simple refinement
         & $0$ & $0$ & $3.9$ & $100$ & $100$
         & $0.1$ & $0$ & $17.8$ & $7.3$ & $0$\\
& Tabu search alone
         & $0$ & $0$ & $3.9$ & $100$ & $100$
         & $0$ & $0$ & $3.9$ & $100$ & $100$\\
& Without PC
         & $4.4$ & $7$ & $13.9$ & $33.3$ & $33.3$
         & $0$ & $0$ & $3.9$ & $100$ & $100$\\
& LSO-NIO & $0$ & $0$ & $7.8$ & $100$ & $0$
         & $2$ & $3$ & $20.3$ & $0$ & $0$\\  \hline \hline
% \multirow{6}{*}{Neg Visit All}
% & AMLSI  & $0$ & $0$ & $0$ & $100$ & $100$
%          & $0$ & $0$ & $0$ & $100$ & $100$\\
% & Generation step
%          & $0$ & $0$ & $0$ & $100$ & $100$
%          & $0$ & $0$ & $13.1$ & $60$ & $0$\\
% & Simple refinement
%          & $2.5$ & $16.0$ & $3.3$ & $40$ & $40$
%          & $0$ & $0$ & $11.3$ & $93.3$ & $20$\\
% & Tabu search alone
%          & $0$ & $0$ & $0$ & $100$ & $100$
%          & $0$ & $0$ & $0$ & $100$ & $100$\\
% & Without PC
%          & $7.3$ & $18.3$ & $18.1$ & $46.7$ & $26.7$
%          & $1.7$ & $3.1$ & $7.2$ & $86.7$ & $80$\\
% & LSO-NIO & $0$ & $0$& $11.1$ & $100$ & $0$
%          & $21.9$ & $10.8$ & $17.6$ & $93.3$ & $0$\\ \hline \hline
\multirow{6}{*}{Neg Elevator}
& AMLSI  & $0$ & $0$ & $0$ & $100$ & $100$
         & ${0}$ & ${0}$ & ${0}$ & $100$ & $100$\\
& Generation step
         & $0$ & $0$ & $0$ & $100$ & $100$
         & $0$ & $7.1$ & $19$ & $13.3$ & $0$\\
& Simple refinement
         & $0$ & $0$ & $0$ & $100$ & $100$
         & $1.6$ & $0.6$ & $14$ & $6.7$ & $0$\\
& Tabu search alone
         & $0$ & $0$ & $5.3$ & $100$ & $66.7$
         & $0$ & $0$ & $5.3$ & $100$ & $66.7$\\
& Without PC
         & $0.5$ & $0.8$ & $2.4$ & $66.7$ & $66.7$
         & $0$ & $0$ & $0.1$ & $100$ & $100$ \\
& LSO-NIO & $0$ & $0$ & $10.8$ & $100$ & $0$
         & $4.8$ & $4.6$ & $16.9$ & $60$ & $0$\\ \hline \hline
\end{tabular}}
\caption{\label{tab:res_complete}Domain learning results when observations are complete.}
\end{center}
\end{subtable}
% \end{table*}
%
% \begin{table*}[!t]
% \begin{center}
% \resizebox{0.6\textwidth}{!}{
\begin{subtable}{0.9\textwidth}
\begin{center}
\resizebox{0.7\textwidth}{!}{
\begin{tabular}{|c|c||c|c|c|c|c||c|c|c|c|c|}
\hline
\multicolumn{2}{|c||}{Noise} & \multicolumn{5}{|c||}{$0\%$} & \multicolumn{5}{|c||}{$20\%$} \\ \hline
Domain & Algorithm & $E_\rho (\%)$ & $E_\epsilon (\%)$ & $E_\sigma (\%)$ & $Solved (\%)$ & $Acc (\%)$ & $E_\rho (\%)$ & $E_\epsilon (\%)$ & $E_\sigma (\%)$ & $Solved (\%)$ & $Acc (\%)$ \\ \hline\hline
\multirow{5}{*}{Blocksworld}
& AMLSI  & $0$ & $0$ & $0$ & $100$ & $100$
         & $2.1$ & $1.5$ & $1.2$ & $76.3$ & $76.3$ \\
& Generation step
         & $0$ & $0$ & $19.1$ & $0$ & $0$
         & $0$ & $0$ & $35.8$ & $0$ & $0$\\
& Simple refinement
         & $0$ & $0$ & $0$ & $100$ & $100$
         & $0$ & $0$ & $28.7$ & $20$ & $0$\\
& Tabu search alone
         & $2.34$ & $4.5$ & $12.3$ & $61.7$ & $9$
         & $2.4$ & $4.7$ & $13.3$ & $67$ & $10.3$ \\
& Without PC
          & $19.2$ & $29.8$ & $22.1$ & $27.7$ & $27.7$
          & $13.3$ & $21.8$ & $16.7$ & $60$ & $36.3$ \\
& LSO-NIO& $0$ & $0$& $28.1$ & $26.7$ & $0$
         & $15.9$ & $18.8$ & $33.4$ & $40.3$ & $0$ \\ \hline \hline
\multirow{6}{*}{Gripper}
& AMLSI  & $0$ & $0$ & $0$ & $100$ & $100$
         & $0$ & $0$ & $0$ & $100$ & $100$ \\
& Generation step
         & $0$ & $0$ & $0.9$ & $86.7$ & $86.7$
         & $0$ & $0$ & $37.6$ & $0.7$ & $0$\\
& Simple refinement
         & $0$ & $0$ & $0$ & $100$ & $100$
         & $0$ & $0$ & $28.5$ & $13.3$ & $0$\\
& Tabu search alone
         & $0$ & $0$ & $0$ & $100$ & $100$
         & $0$ & $0$ & $0$ & $100$ & $100$\\
& Without PC
         & $0$ & $0$ & $0$ & $100$ & $100$
         & $0$ & $0$ & $0$ & $100$ & $100$ \\
& LSO-NIO& $0$ & $0$ & $30$ & $33.3$ & $0$
         & $7.7$ & $5$ & $32.2$ & $13.3$ & $0$ \\ \hline \hline
% \multirow{6}{*}{Hanoi}
% & AMLSI  & $0$ & $0$ & $0.9$ & $100$ & $100$
%          & ${0.5}$ & ${0.9}$ & ${2}$ & $93$ & $86.7$ \\
% & Generation step
%          & $0$ & $0$ & $23.6$ & $0$ & $0$
%          & $0$ & $0$ & $38$ & $4.3$ & $0$\\
% & Simple refinement
%          & $0$ & $0$ & $0.9$ & $100$ & $100$
%          & $3.4$ & $0$ & $35$ & $0.7$ & $0$\\
% & Tabu search alone
%          & $3.7$ & $6.6$ & $18.8$ & $53.3$ & $6.7$
%          & $4.3$ & $6$ & $19.7$ & $49.3$ & $6.7$\\
% & Without PC
%          & $15.6$ & $30.1$ & $21.7$ & $8.3$ & $6.7$
%          & $3.4$ & $7$ & $14$ & $68$ & $26.7$ \\
% & LSO-NIO& $0$ & $0$ & $29.9$ & $12.7$ & $0$
%          & $1.8$ & $5$ & $30.9$ & $6$ & $0$\\ \hline \hline
% \multirow{6}{*}{N-Puzzle}
% & AMLSI  & $0$ & $0$ & $5.6$ & $100$ & $100$
%          & $0$ & $0$ & $5.6$ & $100$ & $100$ \\
% & Generation step
%          & $0$ & $0$ & $27$ & $1.3$ & $0$
%          & $0$ & $0$ & $38.9$ & $0$ & $0$\\
% & Simple refinement
%          & $0$ & $0$ & $5.6$ & $100$ & $100$
%          & $0$ & $0$ & $38.9$ & $0$ & $0$\\
% & Tabu search alone
%          & $0$ & $0$ & $5.6$ & $100$ & $100$
%          & $0$ & $0$ & $5.6$ & $100$ & $100$\\
% & Without PC
%          & $1.1$ & $1.7$ & $6.7$ & $93.3$ & $93.3$
%          & $0$ & $0$ & $5.6$ & $100$ & $100$ \\
% & LSO-NIO& $0$ & $0$ & $28.1$ & $14$ & $0.7$
%          & $0$ & $0$ & $27.7$ & $13.3$ & $0.7$\\ \hline \hline
\multirow{6}{*}{Peg Solitaire}
& AMLSI  & ${0}$ & $0$ & ${4.9}$ & $99.3$ & $99.3$
         & $2.3$ & $1.9$ & $9.9$ & ${78}$ & $61$ \\
& Generation step
         & ${0}$ & $0$ & $20.6$ & $0$ & $0$
         & ${1.6}$ & $0$ & $23.7$ & $0$ & $0$\\
& Simple refinement
         & ${0}$ & $0$ & $11.4$ & $0.7$ & $0$
         & ${2.1}$ & $0$ & $19.3$ & $0$ & $0$\\
& Tabu search alone
         & $3.3$ & $0$ & $11.2$ & $94.7$ & $88$
         & $6.3$ & $2.7$ & $17.7$ & $59$ & $33.7$\\
& Without PC
         & $7.8$ & $15$ & $13.8$ & $59.3$ & $59.3$
         & $4.5$ & $4.3$ & $12.6$ & $67.3$ & $49$ \\
& LSO-NIO& $0$ & $0$ & $19.4$ & $38.7$ & $0$
         & $12.5$ & $6$ & $24$ & $24$ & $0$\\ \hline \hline
\multirow{6}{*}{Parking}
& AMLSI  & $0$ & $0$ & $4.1$ & $100$ & $100$
         & $0.2$ & $0$ & $4.3$ & $99$ & $99$ \\
& Generation step
         & $0$ & $0$ & $21.2$ & $0$ & $0$
         & $0$ & $0$ & $30.9$ & $0$ & $0$\\
& Simple refinement
         & $0$ & $1$ & $4.2$ & $100$ & $77.3$
         & $0$ & $0$ & $30.7$ & $1.3$ & $0$\\
& Tabu search alone
         & $0$ & $0$ & $8.4$ & $86.7$ & $67.3$
         & $0$ & $0$ & $8.3$ & $86.7$ & $70$\\
& Without PC
         & $29.7$ & $42$ & $41.2$ & $0.7$ & $0.3$
         & $6$ & $5.4$ & $11.8$ & $77.7$ & $56.7$ \\
& LSO-NIO& $0$ & $0$ & $25$ & $39.7$ & $0$
         & $11$ & $37.2$ & $28.3$ & $24.7$ & $0$\\ \hline \hline
\multirow{6}{*}{Zenotravel}
& AMLSI  & $0$ & $0$ & $0$ & $100$ & $100$
         & ${0}$ & ${0}$ & ${0}$ & ${100}$ & $100$ \\
& Generation step
         & $0$ & $0$ & $10$ & $12.6$ & $0$
         & $1.7$ & $3$ & $25.9$ & $19.3$ & $0$\\
& Simple refinement
         & ${0}$ & $0$ & $1$ & $92.7$ & $65.6$
         & $2.5$ & $2.4$ & $206$ & $51.3$ & $0$\\
& Tabu search alone
         & ${0}$ & $0$ & $4.2$ & $100$ & $40$
         & ${0}$ & $0$ & $8.8$ & $100$ & $20$\\
& Without PC
         & $4.9$ & $13.5$ & $9.4$ & $69$ & $33.3$
         & $2.8$ & $8.9$ & $7.5$ & $93.3$ & $66.7$ \\
& LSO-NIO& $0$ & $0$ & $25.4$ & $41.3$ & $0$
         & $10.2$ & $7.9$ & $28.4$ & $13.3$ & $0$\\ \hline \hline
\multirow{6}{*}{Sokoban}
& AMLSI  & $0$ & $0$ & $3.9$ & $100$ & $100$
         & $0$ & ${0}$ & $6.1$ & $99.3$ & $60$ \\
& Generation step
         & $0$ & $0$ & $10.6$ & $26.7$ & $26.7$
         & $0$ & $0$ & $19.2$ & $6.7$ & $0$\\
& Simple refinement
         & $0$ & $0$ & $4.3$ & $86.7$ & $86.7$
         & $0.1$ & $0$ & $17.8$ & $13.3$ & $0$\\
& Tabu search alone
         & $0$ & $0$ & $5.4$ & $94.7$ & $73.3$
         & $0$ & ${0}$ & $5$ & $100$ & $80$\\
& Without PC
         & $2.6$ & $5.7$ & $7.7$ & $73.3$ & $73.3$
         & $0$ & $0$ & $6.5$ & $99.3$ &$53.3$ \\
& LSO-NIO& $0$ & $0$ & $19$ & $20$ & $0$
         & $20$ & $20$ & $24.8$ & $0$ & $0$\\  \hline \hline
% \multirow{6}{*}{Neg Visit All}
% & AMLSI  & $0$ & $0$ & $0$ & $100$ & $100$
%          & $0$ & $0$ & $0$ & $100$ & $100$ \\
% & Generation step
%          & $0$ & $0$ & $4.8$ & $20$ & $1$
%          & $0$ & $3.4$ & $13.5$ & $46.7$ & $0$\\
% & Simple refinement
%          & $3$ & $28.4$ & $6.7$ & $26.7$ & $1.3$
%          & $1.4$ & $8.6$ & $12.8$ & $66.7$ & $6.7$\\
% & Tabu search alone
%          & $0$ & $0$ & $0$ & $100$ & $100$
%          & $0$ & $0$ & $0$ & $100$ & $100$ \\
% & Without PC
%          & $8$ & $16.1$ & $17.6$ & $26.7$ & $26.7$
%          & $1.6$ & $1.9$ & $4.4$ & $86.7$ & $86.7$\\
% & LSO-NIO& $0$ & $0$ & $15.2$ & $100$ & $0$
%          & $21.4$ & $14.5$ & $19.3$ & $73.3$ & $0$\\ \hline \hline
\multirow{6}{*}{Neg Elevator}
& AMLSI  & $0$ & $0$ & $0$ & $100$ & $100$
         & $0$ & $0$ & $0$ & $100$ & $100$ \\
& Generation step
         & $0$ & $0$ & $7.4$ & $6.7$ & $0$
         & $0$ & $0$ & $17.2$ & $33.3$ & $0$\\
& Simple refinement
         & $0$ & $0$ & $2$ & $46.7$ & $46.7$
         & $0.2$ & $0$ & $14.2$ & $33.3$ & $0$\\
& Tabu search alone
         & $0$ & $0$ & $5.3$ & $100$ & $66.7$
         & $0$ & $0$ & $3.2$ & $100$ & $80$\\
& Without PC
         & $0.4$ & $0.5$ & $1.8$ & $73.3$ & $73.3$
         & $0$ & $0$ & $0.9$ & $93.3$ & $93.3$ \\
& LSO-NIO & $0$ & $0$ & $19.5$ & $53.3$ & $0$
         & $6.8$ & $7.4$ & $21.7$ & $20$ & $0$\\ \hline \hline
\end{tabular}}
\caption{\label{tab:res_partial}Domain learning results when observations are partial ($25\%$).}
\end{center}
\end{subtable}
\caption{Domain learning results on 7 IPC domains when observations are complete. AMLSI performance is measured in terms of error rates for preconditions (resp. effects) $E_\rho$ (resp. $E_\epsilon$), syntactical error $E_{\sigma}$, the rate of solved problems $Solved$ and accuracy $Acc$. AMLSI's performance are compared with LSO-NIO's performance with the same experimental setup. We report the averages computed over $15$ runs ($5$ runs over $3$ different initial states).}
\end{table*}

%% file: parts/conclusion.tex
\section{Conclusion}
In this paper we present AMLSI, a novel algorithm to learn PDDL domains. We assume that it is possible to query a system to model and to collect partial and noisy observations. AMLSI is composed of four steps. The first step consists in building two training sets of feasible and infeasible action sequences. In the second step, AMLSI induces a regular grammar. The third step is the generation of the PDDL operators, and the last step refines the generated operators. Our experimental results show that AMLSI successfully learns PDDL domains with high levels of noise and incomplete observations and outperforms baseline algorithm.

The performance of AMLSI depends a lot on the induced regular grammar. If the grammar is too complex, or too simple, AMLSI becomes more sensitive to noise and partial observations. As the complexity of the grammar depends on the initial states, future works will focus on the selection of the initial states. Moreover, it should be possible to bias the training set generation in order to obtain the best possible grammar while minimizing the queries. Finally, AMLSI will be extended in order to learn more expressive PDDL operators with disjunctive preconditions, conditional effects and numerical functions.

%% file: parts/ack.tex
\section*{Acknowledgements}
This research is supported by the French National Research Agency under the "Investissements d’avenir” program (ANR-15-IDEX-02) through the Cross Disciplinary Program CIRCULAR.